\documentclass[11pt]{article}
\usepackage{enumitem}
\usepackage{polyglossia}
\setmainlanguage{english}
\setotherlanguage{hindi}
\setotherlanguage{bengali}

\newfontfamily\devanagarifont[Script=Devanagari, Path=fonts/, Extension=.ttf, UprightFont=LohitDevanagari-Regular]{LohitDevanagari}

\newfontfamily\bengalifont[Script=Bengali, Path=fonts/, Extension=.ttf, UprightFont=NotoSansBengali-Regular]{NotoSansBengali}

\usepackage[preprint]{acl}

\usepackage{times}
\usepackage{latexsym}

\usepackage[T1]{fontenc}

\usepackage[utf8]{inputenc}

\usepackage{microtype}

\usepackage{inconsolata}

\usepackage{graphicx}

\title{\centering
IndicTalk: A Large-Scale Persona-Based Multilingual Conversational Corpus for Indic Languages
}

\author{Sahil Deepak Gawande \\
  Lingo Research Group \\  
  IIT Gandhinagar \\
  \texttt{sahil.gawande@iitgn.ac.in} \\\And
  Mayank Singh \\
  Lingo Research Group \\ 
  IIT Gandhinagar \\
  \texttt{singh.mayank@iitgn.ac.in} \\}

\usepackage{amsmath}
\usepackage{array}
\usepackage{booktabs}
\usepackage{tabularx}
\usepackage{fontspec}
\usepackage{polyglossia}
\usepackage{multirow}
\usepackage{hyperref}
\newcommand{\IndicTalk}{\textsc{IndicTalk}}

\begin{document}
\maketitle
\begin{abstract}
Large Language Models (LLMs) have transformed conversational AI, yet high-quality multilingual code-mixed dialogue resources remain scarce, particularly for Indic languages where speakers naturally alternate between English and their native language in both native-script and Romanized forms. We present \IndicTalk, one of the largest multilingual Indic code-mixed conversational corpora, comprising over 13,28,604 event-grounded multi-turn conversations across 18 language varieties covering 9 Indic languages. The corpus is generated through a fully automated pipeline that combines real-world news grounding, persona-conditioned dialogue generation using multilingual LLMs, and automatic quality validation. Extensive linguistic, automatic, and human evaluations demonstrate that \IndicTalk{} produces fluent, coherent, and naturally code-mixed conversations across both script variants. We will release \IndicTalk{} to support the development and evaluation of multilingual conversational AI for underrepresented Indic languages. Dataset is available at Hugging Face: \href{https://huggingface.co/datasets/LingoIITGN/IndicTalk}{IndicTalk Dataset}
\end{abstract}

\section{Introduction}
\label{sec:intro}

Large Language Models (LLMs) have transformed conversational AI, enabling increasingly natural and capable dialogue systems. Their success, however, depends critically on the availability of large-scale, high-quality conversational corpora. While English benefits from abundant dialogue datasets, comparable resources remain scarce for multilingual and code-mixed settings. This gap is particularly evident for Indic languages, where bilingual speakers routinely alternate between English and their native language within the same conversation, using both native scripts and Romanized writing in informal digital communication. Developing conversational systems that serve these multilingual users therefore requires dialogue corpora that accurately capture natural code-mixing patterns across multiple turns.

Although code-mixed NLP has received growing attention, existing Indic datasets primarily target discriminative tasks such as sentiment analysis, hate speech detection, part-of-speech tagging, and natural language inference. Conversational resources are comparatively rare and are typically limited to Hinglish, small-scale collections, or manually curated datasets. To the best of our knowledge, no publicly available resource provides large-scale, event-grounded, multi-turn conversations across multiple Indic code-mixed language varieties under a unified generation framework. This lack of conversational data has become a major bottleneck for training, instruction tuning, and evaluating multilingual conversational LLMs.

To address this gap, we introduce \IndicTalk, the largest multilingual Indic code-mixed conversational corpora to date. The corpus comprises over 13,28,604 event-grounded multi-turn conversations spanning 18 language varieties across 9 Indic languages, covering both native-script and Romanized code-mixed variants paired with English. Each conversation consists of 6--8 dialogue turns generated through a fully automated pipeline that combines real-world news grounding, persona-conditioned dialogue generation using multilingual LLMs, and automatic quality validation, enabling scalable corpus construction without manual annotation.

We perform extensive automatic and human evaluation to assess both conversational quality and code-mixing characteristics. Our evaluation includes established code-mixing metrics, fluency analysis for native-script variants, LLM-as-a-Judge evaluation using Gemini-2.5-Flash~\cite{comanici2025gemini25pushingfrontier} and GPT-OSS-120B~\cite{openai2025gptoss120bgptoss20bmodel}, and human evaluation by native or highly proficient speakers. Results consistently demonstrate that \IndicTalk{} contains fluent, coherent, and naturally code-mixed conversations across all supported language varieties.

Our primary contributions are as follows:

\begin{enumerate}[noitemsep,nolistsep]
    \item We introduce \IndicTalk, the largest multilingual Indic code-mixed conversational corpora, comprising over 13,28,604 event-grounded multi-turn conversations across 18 language varieties covering 9 Indic languages.
    \item We propose a fully automated pipeline for generating multilingual code-mixed conversations that combines real-world news grounding, persona-conditioned dialogue generation, multilingual LLMs, and automatic quality validation.
    \item We provide a comprehensive evaluation of the corpus using linguistic code-mixing metrics, automatic fluency measures, LLM-as-a-Judge assessment, and human evaluation, demonstrating the quality and naturalness of the generated conversations.
\end{enumerate}

\section{Related Work}
\label{sec:rel_work}

Existing work on code-mixed NLP has primarily focused on building datasets for downstream understanding tasks, while comparatively little attention has been devoted to large-scale conversational resources for generative dialogue modeling. We briefly review prior work in three directions: (i) code-mixed conversational datasets, (ii) large-scale Indic conversational resources, and (iii) how our work differs from existing efforts.

\subsection{Code-Mixed Conversational Datasets}

Most existing code-mixed resources for Indic languages are designed for discriminative NLP tasks such as sentiment analysis, hate speech detection, named entity recognition, and natural language inference \cite{chakravarthi-etal-2020-sentiment,srivastava-singh-2020-phinc}. Although valuable, these datasets consist primarily of isolated sentences or social media posts and are not suitable for training conversational LLMs.

Only a handful of datasets focus on code-mixed dialogue. Banerjee et al.~\cite{banerjee-etal-2018-dataset} extended the DSTC2 task-oriented dialogue corpus to four Indic language pairs through human translation. GupShup \cite{mehnaz-etal-2021-gupshup} introduced a Hinglish conversational summarization dataset by translating the SAMSum corpus, while KCM \cite{SINGH2022108900} constructed knowledge-grounded code-mixed dialogues from the Holl-E dataset. More recently, \cite{singh2025sampleefficientlanguagemodelhinglish} demonstrated synthetic Hinglish dialogue generation using LLMs. However, these datasets are restricted to Hinglish or a small number of languages, rely heavily on translation of existing conversations, or target specific domains such as task-oriented or knowledge-grounded dialogue.

\subsection{Indic Conversational Resources}

Large-scale conversational corpora have recently become available for monolingual Indic languages. IndicDialogue \cite{ARNOB2024110690} collects subtitle-based conversations across ten Indic languages, while IndicLLMSuite \cite{khan2024indicllmsuiteblueprintcreatingpretraining} provides large-scale multilingual pre-training and instruction-tuning resources. Although these datasets significantly advance Indic LLM development, they contain predominantly monolingual conversations and do not model natural code-mixing.

\subsection{Our Contribution}

Our work differs from existing efforts along three important dimensions. First, unlike prior resources that are largely limited to Hinglish or a few language pairs, we introduce one of the largest multilingual Indic code-mixed conversational corpora, spanning 18 language varieties across 9 Indic languages in both native-script and Romanized forms. Second, rather than translating or manually curating conversations, we propose a fully automated pipeline that generates event-grounded, persona-conditioned multi-turn dialogues using multilingual LLMs, enabling scalable corpus creation without manual annotation. Finally, we conduct comprehensive automatic and human evaluation, including linguistic code-mixing analysis, LLM-as-a-Judge assessment, and human evaluation, demonstrating the quality and naturalness of the generated conversations.

\section{Methodology}
\label{sec:methodology}
Given a collection of source documents
$\mathcal{D}=\{d_1,d_2,\ldots,d_n\}$,
our objective is to automatically construct a multilingual code-mixed conversational corpus
$\mathcal{C}$ that consists of multi-turn dialogues grounded in real-world events. Here,
$|\mathcal{D}|=n$ denotes the total number of source documents. For each document $d_i$, the pipeline first derives a language-independent semantic representation and subsequently generates conversations independently for each target language variety. Formally,

\[
d_i
\;\xrightarrow{\;\Phi_s\;}\;
s_i
\;\xrightarrow{\;\Phi_g^{(l)}\;}\;
c_i^{(l)}
\;\xrightarrow{\;\Phi_v\;}\;
\hat{c}_i^{(l)},
\]

where $\Phi_s$ denotes the summarization model that generates a language-independent factual summary $s_i$ from the source document $d_i$, $l\in\mathcal{L}$ indexes a target language variety, $\Phi_g^{(l)}$ denotes the dialogue generation model for language variety $l$, $c_i^{(l)}$ is the generated conversation before quality filtering, $\Phi_v$ denotes the automatic validation module, and $\hat{c}_i^{(l)}$ is the validated conversation retained in the final corpus.

The final corpus is

\[
\mathcal{C}
=
\left\{
\hat{c}_i^{(l)}
\;\middle|\;
d_i\in\mathcal{D},\;
l\in\mathcal{L}
\right\},
\]

where $\mathcal{L}$ denotes the set of 18 Indic-English language varieties.

\subsection{Source Document Processing}

The source corpus consists of a collection of publicly available news articles and blog posts covering diverse domains, including \textit{current affairs} and \textit{politics}, \textit{international affairs}, \textit{technology}, \textit{business}, \textit{sports}, \textit{health}, \textit{entertainment}, \textit{science}, \textit{culture}, \textit{education}, and \textit{lifestyle}. 
For each document, the primary textual content is extracted after removing boilerplate elements such as navigation menus, advertisements, scripts, and duplicated template content. Documents with insufficient textual content or unsuccessful extraction are discarded to maintain corpus quality. Table~\ref{tab:category_distribution} presents the composition of the source corpus.  Additional implementation details on source document extraction, preprocessing, and data cleaning are provided in Appendix~\ref{data_cleaning}.

\begin{table}[h]
\centering
\begin{tabular}{lr}
\toprule
\textbf{Category} & \textbf{\# Articles} \\
\midrule
Current Affairs \& Politics  & 11,845 \\
International Affairs        & 10,826 \\
Lifestyle                    & 6,465 \\
Health                       & 6,056 \\
Technology                   & 5,366 \\
Entertainment                & 5,056 \\
Business                     & 4,690 \\
Education                    & 3,476 \\
Science                      & 3,362 \\
Sports                       & 2,442 \\
Culture                      & 1,562 \\
Other                        & 80,907 \\
\bottomrule
\end{tabular}
\caption{Composition of the source corpus.}
\label{tab:category_distribution}
\end{table}

\subsection{Language-Independent Semantic Representation}

Generating conversations directly from raw documents often introduces unnecessary variability due to differences in document length, writing style, and formatting. Longer documents, in particular, tend to bias LLMs towards producing disproportionately longer or more detailed conversations. We therefore first transform each document into a language-independent semantic representation, which normalizes the input, reduces document-length bias, and enables all language variants to be generated from the same factual content while minimizing stylistic variation.

Given a document $d_i$, the summarization model produces a concise English summary

\[
s_i=\Phi_s(d_i),
\]

which serves as the common semantic representation for dialogue generation across all target languages. Using a single English representation avoids generating and maintaining separate summaries for each language, ensuring that every conversation remains grounded in identical factual content while allowing language-specific realization during dialogue generation. We constrain the summary to 6--8 factual sentences to balance contextual completeness with generation stability, providing sufficient information for generating coherent 6--8-turn conversations while avoiding unnecessary details that can lead to topic drift. The summarization prompt preserves named entities, numerical values, temporal expressions, and locations while discouraging unsupported information.  

We compare multiple open-source multilingual models through an arena-style blind human evaluation, where annotators independently compare summaries generated by competing models. As shown in Table~\ref{tab:sarvam_eval}, Sarvam-M~\cite{sarvam} consistently outperforms Llama~7B~\cite{llama} and Gemma~7B~\cite{gemma}, achieving the highest Elo rating and win rate. We therefore select Sarvam-M~\cite{sarvam} as the summarization model for the remainder of the pipeline. Additional details of the evaluation protocol are provided in Appendix~\ref{Sarvam_eval}, while the summarization prompt is included in Appendix~\ref{app:prompts}.

\begin{table}[t]
\centering
\begin{tabular}{lccc}
\toprule
\textbf{Model} & \textbf{Elo} & \textbf{W-L-T} & \textbf{Battles} \\
\midrule
\textbf{Sarvam-M} & 1861.5 & 267-18-36 & 321 \\
Llama 7B & 1582.1 & 162-104-36 & 302 \\
Gemma 7B & 1056.4 & 11-318-2 & 331 \\
\bottomrule
\end{tabular}
\caption{Arena-style blind human evaluation of three summarization models. Sarvam-M achieves the highest Elo rating and win-loss record.}
\label{tab:sarvam_eval}
\end{table}

\begin{table*}[t]
\centering

\begin{tabular}{p{2.5cm}p{4.5cm}p{7cm}}
\toprule
\textbf{Category} & \textbf{Representative Role Pairs} & \textbf{Interaction Style} \\
\midrule
Friends &
Friend--Friend &
Informal conversations between friends. \\

Family Members &
Mother--Son, Father--Daughter, Siblings, Grandparent--Grandchild, \emph{etc.} &
Randomly sampled interaction styles including supportive, curious, advisory, skeptical, optimistic, and debate. \\

Colleagues &
Manager--Employee, Mentor--Mentee, Peer Colleagues, Team Lead--Junior, \emph{etc.} &
Professional workplace discussions between hierarchical and peer-level roles. \\

Experts &
Expert--Expert &
Technical discussions involving analysis and domain knowledge. \\

Student--Teacher &
Student--Teacher &
Educational conversations involve explanation, questioning, and clarification. \\

\bottomrule
\end{tabular}
\caption{Persona configurations used for dialogue generation.}
\label{tab:personas}
\end{table*}
\begin{table*}[t]
\centering
\renewcommand{\arraystretch}{1.2}

\begin{tabular}{p{2cm}p{13cm}}
\toprule
\textbf{Variant} & \textbf{Example Conversation} \\
\midrule

\textbf{Native-Script Code-Mixing} &
\textbf{Speaker 1:}
{\bengalifont গতকাল আমি রাস্তায় গিয়ে একটা সাইকেলিস্টের সামনে হঠাৎই গাড়ি থেমে গেল, আমার মনটা} literally froze {\bengalifont হয়ে গেল} {\bengalifont । আমি জানি সেটা} AEB system {\bengalifont এর} thanks, {\bengalifont কারণ} sensor {\bengalifont গুলো ৪৫} mph {\bengalifont পর্যন্ত} non-moving target detect {\bengalifont করতে পারে, আর আমি} safety feeling {\bengalifont পেয়ে বেশ} relaxed {\bengalifont লাগলো} {\bengalifont । যদিও} automakers \ldots

\medskip

\textbf{Speaker 2:}
{\bengalifont ওহ্, তুমি হঠাৎই গাড়ি থামতে দেখলেই আমি} totally shocked {\bengalifont হলাম!} AEB sensor {\bengalifont গুলো ৪৫} mph {\bengalifont পর্যন্ত} non-moving target detect {\bengalifont করতে পারে, কিন্তু কি তারা সত্যি ৩৬০} lives yearly {\bengalifont বাঁচাতে পারে?}regulators {\bengalifont ৬২} mph \ldots
\\[1ex]

\textbf{Romanized Code-Mixing} &

\textbf{Speaker 1:}
Kal amar choto bhai-er birthday-te sobai baby photo dekhte dekhte, amar mon ektu nostalgic hoye galo, tokhon ei 5-week-old twins er murder suspicion er news shune amar khub shock \ldots

\medskip

\textbf{Speaker 2:}
Ami ekta shishu hospital-e volunteer korar somoy ekta choto baby ke niye onek emotional moment dekhlam, tar mayer dukkho dekhte amar mon bhenge gelo. Tokhon ami bujhte parlam je shishuder safety aar \ldots
\\
\bottomrule
\end{tabular}
\caption{Representative examples of native-script and romanized code-mixed conversations.}
\label{tab:code_mixing_variants}
\renewcommand{\arraystretch}{1.0}
\end{table*}

\subsection{Persona Specification}

To encourage conversational diversity, each dialogue is conditioned on a pair of personas representing common real-world interactions. We define five persona categories: \emph{Friends}, \emph{Family Members}, \emph{Colleagues}, \emph{Experts}, and \emph{Student--Teacher}. While the \emph{Friends}, \emph{Experts}, and \emph{Student--Teacher} categories use fixed role pairs, the remaining categories randomly sample role pairs and interaction styles from predefined subcategories. The selected personas remain fixed throughout the conversation to ensure consistent context. Table~\ref{tab:personas} summarizes the persona configurations.

\subsection{Multilingual Dialogue Generation}

For each semantic representation $s_i$, the dialogue generation model
$\Phi_g^{(l)}$ independently generates a conversation
$c_i^{(l)}$ for every target language variety
$l\in\mathcal{L}$ using a multilingual LLM under a self-play framework. The generated conversation before validation is $
c_i^{(l)}
=
\{u_1,u_2,\ldots,u_T\},
$ where $T$ denotes the number of dialogue turns. Each utterance is generated autoregressively as $
u_t
=
\Phi_g^{(l)}
\left(
s_i,
P,
u_{<t}
\right),
$ where $P$ denotes the persona specification and $u_{<t}$ represents the dialogue history up to turn $t$.

For each Indic language, we generate two complementary code-mixed variants: \textbf{Native-Script Code-Mixing}, where the Indic language is written in its native script and English remains in Roman script, and \textbf{Romanized Code-Mixing}, where both languages are written entirely in Roman script. Representative examples of both variants are shown in Table~\ref{tab:code_mixing_variants}.

\subsection{Automatic Conversation Validation}

Despite explicit prompting, multilingual LLMs occasionally generate monolingual or weakly code-mixed conversations. We therefore apply an automatic validation module, $\Phi_v$, to filter generated conversations before constructing the final corpus. A conversation is retained only if it satisfies the following criteria:

\begin{enumerate}[nosep, noitemsep]
    \item \textbf{Script Validation:} For native-script variants, each utterance must contain at least one token in the target Indic script and one token in the Roman script.

    \item \textbf{Code-Mixing Threshold:} For native-script variants, every utterance must satisfy a minimum Code-Mixing Index (CMI)~\cite{das-gamback-2014-identifying}, i.e., $\mathrm{CMI}\geq\tau_{\text{CMI}}$.

    \item \textbf{Minimum Length:} Each utterance must contain at least $\tau_{\text{len}}$ tokens.
\end{enumerate}

For Romanized variants, script validation simply verifies the use of Roman characters, while the CMI constraint is omitted because both the Indic language and English share the same script, making token-level language identification unreliable and artificially penalizing natural Romanized code-mixing~\cite{benton-etal-2025-improving}. The quality of these conversations is assessed instead through the human evaluation described in Section~\ref{section_4}.

\subsection{Model Selection for Dialogue Generation}
We experimented with several state-of-the-art instruction-tuned language models for multilingual dialogue generation, including GPT-OSS-120B, Qwen2.5-72B, Gemma 4-31B, and their smaller variants. During the iterative development of the generation pipeline, we observed that GPT-OSS-120B consistently produced dialogues that more reliably satisfied our downstream validation criteria, including persona consistency, multilingual fidelity, and structural correctness. In contrast, outputs from the other models more frequently failed one or more validation stages, requiring substantially more regeneration attempts. Based on these empirical observations, GPT-OSS-120B was selected as the primary dialogue generation model for the final dataset. Reference Generation model is present in Appendix~\ref{comparison}. 

\section{Dataset Analysis}
\label{section_4}

We analyze \IndicTalk{} from three perspectives. First, we summarize the corpus composition and scale. Second, we characterize its code-mixing properties using established linguistic metrics. Finally, we evaluate conversation quality through automatic metrics, human evaluation, and LLM-as-a-Judge assessment.

\subsection{Dataset Statistics}

Tables~\ref{tab:dataset_statistics} and~\ref{tab:size of dataset} summarize the statistics of \IndicTalk{}. The corpus comprises over 13,28,604 multilingual code-mixed conversations generated from approximately 1,42,053 news articles and blog posts spanning 12 domains. It covers 9 Indic languages, each represented in both native-script and Romanized code-mixed variants, resulting in \textbf{18} language varieties. Each conversation contains an average of \textbf{7.55} dialogue turns and is grounded in the same source document across language variants, enabling controlled cross-lingual and cross-script comparisons while preserving the underlying factual content.

\begin{table}[t]
\centering

\begin{tabular}{lr}
\toprule
\textbf{Statistic} & \textbf{Value} \\
\midrule
News Sources / Blogs & 1,42,053 \\
Domains Covered & 12 \\
Language Variants & 9 \\
Script Variants & 2 \\
Persona Categories & 5 \\
Total Conversations & 13,28,604 \\
Average Turns/conv & 7.55 \\
Standard Deviation & 0.83 \\
Total turns & 10,691,164 \\

\bottomrule
\end{tabular}
\caption{Overall statistics of the proposed multilingual code-mixed conversational corpus.}\label{tab:dataset_statistics}
\end{table}

\begin{table}[t]
\centering
\caption{Total Corpus size along with language-wise distribution, CM- Code-Mixing}
\label{tab:size of dataset}

\begin{tabular}{lrr}
\toprule
\textbf{Languages} &
\textbf{Native} &
\textbf{Romanized} \\
\midrule
Bengali    & 100,168  & 45,020  \\
Gujarati   & 105,004 &  45,012 \\
Hindi      & 101,303 & 45,014  \\
Kannada    & 104,988 & 45,012  \\
Malayalam  & 104,273 & 45,010  \\
Marathi    & 103,464 & 45,021 \\
Odia       & 101819 & 45,005  \\
Tamil      & 100,409 & 45,006  \\
Telugu     & 102,069 & 45,007 \\
\midrule
\textbf{Total} &
\textbf{9,23,497} &
\textbf{4,05,107} \\
\bottomrule
\end{tabular}
\end{table}

\subsection{Code-Mixing Characteristics}

We characterize the code-mixing properties of \IndicTalk{} using four complementary metrics: Code-Mixing Index (CMI)~\cite{das-gamback-2014-identifying}, Switch Point Fraction (SPF)~\cite{pratapa-etal-2018-language}, Integration Index (I-index)~\cite{guzman17_interspeech}, and Multilingual Index (M-index)~\cite{Bernett}. 

Table~\ref{tab:automatic_analysis} summarizes the corpus-level statistics. Across the native-script variants, the corpus achieves an average CMI of 37.81, indicating a high degree of code-mixing according to the formulation of Das and Gambäck~\cite{das-gamback-2014-identifying}. The average SPF (0.438) and I-index (0.437) indicate that language switches occur frequently throughout the conversations rather than being confined to isolated insertions~\cite{pratapa-etal-2018-language,guzman17_interspeech}. Likewise, the average M-index of 0.865 suggests a well-balanced contribution of both languages across the corpus, rather than dominance by a single language~\cite{Bernett}. Telugu and Kannada exhibit the strongest bilingual mixing, while all native-script variants maintain consistently high values across the four metrics.

For the Romanized variants, the average SPF (0.311), I-index (0.311), and M-index (0.508) remain consistently positive across all languages, confirming that meaningful code-mixing is preserved despite the shared writing system. The lower metric scores compared to native-script variants is expected, as automatic language identification becomes less reliable when both languages are written in Roman script, leading to conservative estimates of language balance. Overall, these statistics demonstrate that \IndicTalk{} captures frequent language alternation and balanced bilingual usage across both writing conventions.

\begin{table*}[t]
\centering
\begin{tabular}{lccccc|cccc}
\toprule
\multirow{2}{*}{\textbf{Language}} &
\multicolumn{5}{c|}{\textbf{Native Script}} &
\multicolumn{4}{c}{\textbf{Romanized}} \\
\cmidrule(lr){2-6}
\cmidrule(lr){7-10}
& \textbf{CMI}
& \textbf{SPF}
& \textbf{I-index}
& \textbf{M-index}
& \textbf{PPPL}\footnote{PPPL is reported only for native-script variants. Due to the absence of standardized orthography and the high spelling variability in Romanized Indic text, pseudo-perplexity scores are not directly comparable across Romanized language varieties.}
& \textbf{CMI}
& \textbf{SPF}
& \textbf{I-index}
& \textbf{M-index} \\
\midrule
Bengali   & 33.05 & 0.389 & 0.387 & 0.786 & 14.87 & 20.13 & 0.322 & 0.321 & 0.474 \\
Gujarati  & 42.56 & 0.432 & 0.433 & 0.943 & 18.52 & 13.48 & 0.235 & 0.235 & 0.305 \\
Hindi     & 32.64 & 0.394 & 0.393 & 0.777 & 14.36 & 10.30 & 0.136 & 0.139 & 0.535 \\
Kannada   & 43.15 & 0.482 & 0.480 & 0.949 & 10.33 & 26.81 & 0.425 & 0.425 & 0.642 \\
Malayalam & 38.66 & 0.456 & 0.455 & 0.886 &  9.14 & 35.17 & 0.454 & 0.454 & 0.831 \\
Marathi   & 36.74 & 0.448 & 0.447 & 0.859 & 14.69 & 13.66 & 0.236 & 0.236 & 0.309 \\
Odia      & 32.77 & 0.413 & 0.411 & 0.781 & 14.35 & 28.42 & 0.423 & 0.423 & 0.683 \\
Tamil     & 35.17 & 0.445 & 0.443 & 0.827 & 13.93 & 10.81 & 0.198 & 0.197 & 0.239 \\
Telugu    & 45.52 & 0.485 & 0.485 & 0.976 & 12.00 & 23.25 & 0.369 & 0.369 & 0.554 \\
\midrule
\textbf{Average}
& \textbf{37.81}
& \textbf{0.438}
& \textbf{0.437}
& \textbf{0.865}
& \textbf{13.58}
& \textbf{20.23}
& \textbf{0.311}
& \textbf{0.311}
& \textbf{0.508} \\
\bottomrule
\end{tabular}
\caption{Automatic analysis of the proposed multilingual code-mixed conversational corpus. CMI measures the degree of code-mixing, SPF measures language switching frequency, I-index captures language integration, M-index quantifies language balance, and PPPL measures conversational fluency computed using mBERT.} \label{tab:automatic_analysis}
\end{table*}

\subsection{Fluency Analysis}
While the code-mixing metrics quantify bilingual structure, they do not directly measure linguistic fluency. We therefore evaluate the native-script variants using Pseudo Perplexity (PPPL) computed with multilingual BERT (mBERT) following the masked language model scoring approach of \citet{salazar-etal-2020-masked}, where lower PPPL indicates greater linguistic fluency.

As shown in Table~\ref{tab:automatic_analysis}, the native-script variants achieve consistently low PPPL scores (9.14--18.52), indicating fluent and well-formed conversations despite frequent code-switching. PPPL is not reported for Romanized variants because the absence of standardized orthography and high transliteration variability make pseudo-perplexity scores less comparable across languages. We instead assess their fluency through the LLM-as-a-Judge evaluation described in Section~\ref{llm-as-judge}.

\subsection{LLM-as-a-Judge Evaluation} 
\label{llm-as-judge}
\begin{table}[h]
\centering
\begin{tabular}{l|cccc}
\toprule
\multirow{2}{*}{\textbf{Language}} &
\multicolumn{2}{c}{\textbf{Native}} &
\multicolumn{2}{c}{\textbf{Romanized}} \\
\cmidrule(lr){2-3} \cmidrule(lr){4-5}
& \textbf{Gemini} & \textbf{GPT}
& \textbf{Gemini} & \textbf{GPT} \\
\midrule
Bengali   & 4.42 & 4.03 & 4.50 & 4.03 \\
Gujarati  & 4.34 & 4.04 & 4.28 & 4.05 \\
Hindi     & 4.79 & 4.27 & 4.79 & 4.27 \\
Kannada   & 4.09 & 4.06 & 4.10 & 3.86 \\
Malayalam & 4.08 & 4.01 & 4.18 & 4.03 \\
Marathi   & 4.54 & 4.22 & 4.23 & 3.94 \\
Odia      & 3.82 & 3.62 & 3.55 & 3.76 \\
Tamil     & 4.12 & 4.05 & 4.18 & 4.03 \\
Telugu    & 4.43 & 4.05 & 4.23 & 4.02 \\
\bottomrule
\end{tabular}
\caption{Mean overall quality scores on a 5-point Likert scale (1 = Very Poor, 5 = Excellent) for native-script and Romanized conversations. Gemini denotes Gemini-2.5-Flash and GPT denotes GPT-OSS-120B. Results are averaged over 500 conversations per language.}
\label{tab:llm_judge}
\end{table}

Following the LLM-as-a-Judge paradigm~\cite{zheng2023judgingllmasajudgemtbenchchatbot,liu-etal-2023-g}, we evaluate a held-out set of 500 conversations per language (9,000 conversations across 18 language varieties) using two independent judge models: Gemini-2.5-Flash\cite{comanici2025gemini25pushingfrontier} and GPT-OSS-120B\cite{openai2025gptoss120bgptoss20bmodel}. Since GPT-OSS-120B is also used for data generation, Gemini-2.5-Flash serves as an independent judge to mitigate potential self-preference bias. Each conversation is rated on five dimensions: \textit{Fluency}, \textit{Coherence}, \textit{Engagement}, \textit{Code-Mixing Naturalness}, and \textit{Overall Quality}, using a 1--5 Likert scale. The complete evaluation prompt used for both judge models is provided in Appendix~\ref{llm_judge_prompt} for reproducibility.

As shown in Table~\ref{tab:llm_judge}, both judges consistently assign high scores across all 18 language varieties, with the majority achieving an overall quality score above 4.0. Native-script and Romanized variants exhibit comparable quality, indicating that the proposed generation pipeline produces fluent, coherent, and naturally code-mixed conversations under both writing conventions. The overall quality rankings produced by the two judges are highly correlated ($\rho=0.91$, $p<0.001$), despite Gemini-2.5-Flash assigning consistently higher absolute scores than GPT-OSS-120B, reflecting differences in judge calibration reported in prior work~\cite{zheng2023judgingllmasajudgemtbenchchatbot}. 

\subsection{Human Evaluation}

\begin{table}[h]
    \centering
    \caption{Mean overall quality scores of Human Evaluation per language variety, evaluated by Human annotators on 10 conversations per language variety}
    \label{tab:Human_eval}
    \begin{tabular}{l|cc}
    \toprule
    \textbf{Languages} & \textbf{Native} & \textbf{Romanized} \\
    \midrule
    Bengali    & 4.25 & 3.4 \\
    Gujarati   & 3.7 & 3.1  \\
    Hindi   & 4.0 & 4.4 \\
    Kannada    & 4.3 & 3.0 \\
    Malayalam  & 3.8 & 3.3 \\
    Marathi    & 4.4  & 3.75 \\
    Odia       & 3.6 & 3.2 \\
    Tamil      & 3.61  & 3.5  \\
    Telugu     & 3.7  & 3.0 \\
\bottomrule
    \end{tabular}
\end{table}

To complement the LLM-as-a-Judge evaluation, we conduct a human evaluation on a held-out set of 10 conversations per language variety (180 conversations in total). Each conversation is independently evaluated by at least 2--3 graduate students who are native or highly proficient speakers of the corresponding language. Following the same protocol as the LLM-as-a-Judge evaluation, annotators rate each conversation on \textit{Fluency}, \textit{Coherence}, \textit{Engagement}, \textit{Code-Mixing Naturalness}, and \textit{Overall Quality} using a 5-point Likert scale. The complete annotation guidelines are provided in Appendix~\ref{human_eval_prompt}.

Table~\ref{tab:Human_eval} reports the mean overall quality scores across languages and script variants. Consistent with the LLM-as-a-Judge evaluation, native-script variants generally receive slightly higher scores than their Romanized counterparts, although both achieve high overall quality across languages. Marathi and Kannada obtain the highest human ratings among native-script variants, while Hindi receives the highest score for Romanized conversations. 

\subsection{Overall Evaluation Summary}
The three evaluation protocols consistently indicate that \IndicTalk{} contains high-quality multilingual code-mixed conversations across both script variants. The majority of language varieties achieve overall quality scores above 4.0 under both LLM judges and above 3.0 under human evaluation, demonstrating that the proposed generation pipeline produces fluent, coherent, and naturally code-mixed conversations at scale. To quantify agreement, we compute Spearman rank correlations across language varieties. The two LLM judges exhibit strong agreement in their relative quality rankings ($\rho=0.745$, $p=0.02$ for native-script and $\rho=0.778$, $p=0.01$ for Romanized variants). Human rankings also show moderate positive correlation with both LLM judges($\rho = 0.559$ with GPT-OSS-120B and $\rho = 0.469$ with Gemini-2.5-Flash), indicating broadly consistent quality assessments despite differences in score calibration.

\paragraph{Implementation Details.\label{implementation}} The complete corpus was generated on 16 NVIDIA H200 GPUs over approximately 10,000 GPU-hours. Models were served through OpenAI-compatible REST APIs and queried via standard HTTP requests with a timeout of 90 seconds per request. Document summarization was performed using Sarvam-M, while dialogue generation employed GPT-OSS-120B. For summarization, we used a low decoding temperature ($T=0.1$) to produce stable factual summaries. Dialogue generation used a temperature of $T=0.6$ for the opening utterance and $T=0.7$ for subsequent turns to encourage conversational diversity while preserving coherence. The maximum generation budget was set to 400 tokens for summarization and 600 tokens per dialogue turn. Thinking mode was disabled for Sarvam-M via the \texttt{chat\_template\_kwargs} parameter.

\section{Conclusion}
We presented \IndicTalk, a large-scale multilingual Indic code-mixed conversational corpus comprising over 13,28,604 event-grounded multi-turn conversations across 18 language varieties covering 9 Indic languages in both native-script and Romanized forms. The corpus is constructed through a fully automated pipeline combining real-world news grounding, persona-conditioned dialogue generation, multilingual LLMs, and automatic quality validation. Extensive automatic and human evaluations demonstrate that \IndicTalk{} contains fluent, coherent, and naturally code-mixed conversations across all supported languages, establishing it as a valuable resource for developing and evaluating multilingual conversational AI for Indic languages.

\section{Future Work}
Future work includes extending \IndicTalk{} to additional Indic languages, dialects, and script variants, as well as incorporating richer personas and interaction scenarios such as customer support and multilingual debates. Another promising direction is to establish standardized benchmarks for downstream tasks, including conversational summarization, sentiment analysis, and intent detection, enabling systematic evaluation and comparison of multilingual conversational models for Indic code-mixed NLP.

\section{Limitations}
Although \IndicTalk{} is generated using a carefully designed pipeline with multiple quality control stages, it remains a synthetic corpus and may not fully capture the pragmatic nuances, dialectal variation, and disfluencies found in naturally occurring bilingual conversations. Since the source documents are primarily drawn from news and blogs, the corpus also exhibits a topical bias toward formal and event-centric discourse. In addition, human evaluation is conducted with 2--3 graduate student annotators per language variety and may not fully reflect the diversity of regional dialects and Romanization conventions. Finally, automatic fluency evaluation is reported only for native-script variants, as pseudo-perplexity is not directly comparable for Romanized Indic text.

\section{Ethical Considerations}
\IndicTalk{} is generated from publicly available news articles and blogs using a fully automated pipeline. We do not release the source documents and remove personally identifiable information (PII) where detected during preprocessing. Human evaluation was conducted with informed consent from volunteer graduate student annotators. All datasets and models used in this work will be publicly available or used in accordance with their respective licenses. We hope that \IndicTalk{} facilitates more inclusive research on multilingual code-mixed conversational AI for underrepresented Indic languages.

\bibliography{custom}

\section*{Appendix}
\appendix
\section{Source Data Extraction and Cleaning}\label{data_cleaning}
For each URL in the source collection, content is fetched using a standard HTTP GET request with a browser-mimicking User-Agent header. HTML pages are parsed using BeautifulSoup, with boilerplate elements including navigation menus, headers, footers, sidebars, scripts, and advertisement blocks removed prior to text extraction. PDF documents are handled using pdfplumber, with text extracted page by page and concatenated.

Documents are discarded if the HTTP request fails or if the extracted text falls below a minimum length threshold of 100 words. The pipeline processes articles in sequential batches over non-overlapping index ranges, allowing multiple instances to run in parallel. Each article is wrapped in fault-tolerant error handling so that individual failures do not interrupt the broader pipeline run.

\section{Summarization}

\subsection{Model Selection and Evaluation}
\label{Sarvam_eval}
We evaluate three multilingual open-source models: Sarvam-M\cite{sarvam}, Llama~7B\cite{llama}, and Gemma~7B\cite{gemma}, using an arena-style blind human evaluation to select the summarization model. The evaluation is performed on source documents sampled from four language tracks: English, Gujarati, Hindi, and Tamil, reflecting the linguistic diversity of the corpus.

For each document, annotators compare summaries generated by two models with their identities hidden. The original article and source text are provided to verify factual accuracy, content coverage, and preservation of named entities. Model identities are revealed only after the vote is cast, ensuring unbiased preference judgments.

A total of 509 pairwise preference votes are collected. Sarvam-M achieves the highest Elo rating (1861.5), consistently outperforming Llama~7B and Gemma~7B across all language tracks (Table~\ref{tab:sarvam_eval}). Based on these results, we use Sarvam-M as the summarization model throughout the pipeline.

\subsection{Article Summarization Prompt}
\label{app:prompts}

The summarization model is prompted to generate a language-independent semantic representation of each source document using the following instruction.

\begin{quote}
\small
\textbf{Instruction:} Summarize the following article in 6--8 concise, factual, and neutral English sentences. Preserve all important named entities, numerical values, dates, temporal expressions, and locations. Include only information explicitly supported by the article. Do not introduce opinions, speculation, or external knowledge. Write the output as a coherent paragraph without bullet points.

\vspace{0.5em}
\textbf{Article:}

\{\textit{Input Article}\}

\vspace{0.5em}
\textbf{Summary:}
\end{quote}

\section{LLM-as-a-Judge Prompt 
\label{llm_judge_prompt}}
\begin{figure}[h]
\small
\begin{quote}
\textbf{Prompt Template for LLM-as-Judge Evaluation}

\medskip

You are an expert linguistic annotator evaluating a synthetically generated \texttt{\{script\_name\}} dialogue between two speakers (Speaker\_1 and Speaker\_2), simulating a conversation between different persons discussing on a topic.

\textit{For Romanized variants only: This is a fully Romanized (Latin script) code-mixed dialogue. The speakers use transliterated native-language words mixed with English entirely in Roman script, with no native script characters. This style reflects how many urban Indians text in their regional languages.}

Evaluate the conversation below on the following five metrics. Score each metric on a scale of 1--5 (1 = very poor, 5 = excellent).

\begin{itemize}
\item \textbf{Fluency}: Whether the dialogue is grammatically correct, natural, and easy to read.
\item \textbf{Coherence}: Whether each utterance logically follows the previous turns and maintains a consistent topic.
\item \textbf{Code-Mixing Naturalness}: Whether switching between English and the native language is natural, contextually appropriate, and resembles real bilingual communication.
\item \textbf{Engagement}: Whether the conversation is interesting, interactive, and resembles a realistic exchange rather than a mechanical question-answer pattern.
\item \textbf{Overall Quality}: A holistic assessment considering fluency, coherence, code-mixing naturalness, and engagement.
\end{itemize}

\textbf{Conversation}

\texttt{\{conversation\}}

Return a valid JSON object containing, for each metric, an integer score and a one-sentence justification.

\end{quote}

\caption{Prompt template used for LLM-as-Judge evaluation. The same prompt was used for Gemini-2.5-Flash and GPT-OSS-120B across all language varieties. For Romanized variants, the italicized instruction was additionally included. Both judges used greedy decoding ($T=0.0$), and invalid outputs were retried up to three times.}
\label{fig:judge_prompt}
\end{figure}

\section{Human Evaluation Prompt}
\label{human_eval_prompt}

\small

\noindent\textbf{Prompt Template for Human Evaluation}

\begin{quote}

You are evaluating a synthetically generated \texttt{\{script\_name\}} dialogue between two speakers (Speaker\_1 and Speaker\_2), simulating a conversation between persons discussing a topic.

\textit{For Romanized variants only: The dialogue is written entirely in Roman script, with no native script characters. The speakers mix transliterated native-language words with English, reflecting how many urban Indians text informally in their regional languages.}

Please read the conversation carefully and rate it on the following five dimensions. Score each dimension on a scale of 1--5 (1 = very poor, 5 = excellent). Be strict in your assessment---penalize grammatical errors, spelling mistakes, unnatural expressions, and implausible code-switching patterns as you would encounter them in real informal digital communication.

\begin{itemize}
    \item \textbf{Fluency}: Is the dialogue grammatically correct and natural? Penalize spelling mistakes, morphologically incorrect word forms, and ungrammatical constructions.
    \item \textbf{Coherence}: Does each turn logically follow from the previous ones? Does the conversation maintain a consistent topic throughout?
    \item \textbf{Code-Mixing Naturalness}: Does the switching between the native language and English feel natural and contextually appropriate? Does it resemble how people actually write informally, or does it feel forced and artificial?
    \item \textbf{Engagement}: Is the conversation interesting and dynamic? Does it feel like a realistic exchange between two people, or does it feel mechanical and repetitive?
    \item \textbf{Overall Quality}: Taking all four dimensions into account, how would you rate the overall quality of this conversation?
\end{itemize}

\textbf{Conversation}

\texttt{\{conversation\}}

For each dimension, provide an integer score from 1 to 5 and a brief justification in one or two sentences explaining your rating.

\end{quote}

\section{Qwen2.5-72B Vs Gemma 4-31B Vs GPT-OSS-120B \label{comparison}}
\noindent\textbf{Example Conversations.} To qualitatively compare the conversational capabilities of different LLMs, we present representative outputs generated by Qwen2.5-72B refer Table~\ref{tab:qwen_example}, Gemma~4~31B refer Table~\ref{tab:gemma_example}, and GPT-OSS-120B refer to Table~\ref{tab:gptoss_example} . Among the three, GPT-OSS-120B consistently produces the most coherent, contextually grounded, and naturally code-mixed conversations, while the other models exhibit relatively weaker grounding or conversational consistency.

\begin{table*}[t]
\centering
\small
\caption{Example Hindi-English code-mixed conversation generated by the proposed pipeline, model-Qwen 2.5-72B.}
\label{tab:qwen_example}
\begin{tabular}{lp{0.82\textwidth}}
\toprule
\textbf{Speaker} & \textbf{Utterance} \\
\midrule

Speaker\_1 &
\texthindi{यह} news \texthindi{थोड़ा} suspicious \texthindi{लग रही है, यानी} Israel \texthindi{की} defense minister \texthindi{ने} responsibility \texthindi{ली है,} but \texthindi{कुछ} details missing \texthindi{हैं।} Larijani's death \texthindi{ने तो इरान की} leadership \texthindi{में एक बड़ा वाक्यांश छोड़ दिया है, और उनकी} retaliation \texthindi{की} threat serious \texthindi{है।} \\

Speaker\_2 &
Absolutely, Larijani \texthindi{की} death \texthindi{ने इरान की} internal dynamics \texthindi{पर बहुत} impact \texthindi{डाला है। मेरे एक} friend \texthindi{की} family \texthindi{इरान में रहती है, उन्होंने बताया था कि} tension \texthindi{काफी} high \texthindi{है, और} people scared \texthindi{हैं कि} what next \texthindi{होगा।} \\

Speaker\_1 &
Larijani \texthindi{की} death \texthindi{के बाद, मेरा} cousin \texthindi{जो} Dubai \texthindi{में} lives, \texthindi{उसने इरान से अपने} family members \texthindi{को बुलाने की} plan \texthindi{बना रखी है, क्योंकि वह सोचता है कि} situation worse \texthindi{हो सकती है। वह चिंतित है कि इरान की} retaliation \texthindi{क्या होगी, और उससे उनकी} safety \texthindi{पर क्या होगा।} \\

Speaker\_2 &
Wow, \texthindi{यह तो} really serious \texthindi{है!} Your cousin's concern understandable \texthindi{है,} situation \texthindi{इतनी} unpredictable \texthindi{हो गई है कि} safety first \texthindi{होनी चाहिए।} \\

Speaker\_1 &
I get where your cousin is coming from, but \texthindi{इरान की} retaliation \texthindi{ज़रूरी है, यह उनका आत्म-सम्मान और सुरक्षा का मुद्दा है। फिर भी, हर किसी को अपनी} safety \texthindi{के बारे में ध्यान रखना चाहिए, लेकिन} panic \texthindi{मत करना चाहिए।} \\

Speaker\_2 &
I see your point, but panic \texthindi{नहीं करना चाहिए, इससे स्थिति} control \texthindi{से बाहर निकल सकती है।} Safety measures \texthindi{लेना ज़रूरी है, लेकिन} calm \texthindi{रहना और सोच-समझकर} action \texthindi{लेना बेहतर होगा।} \\

\bottomrule
\end{tabular}
\end{table*}

\begin{table*}[t]
\centering
\small
\caption{Example Hindi-English code-mixed conversation generated by the proposed pipeline,model-Gemma 4-31B.}
\label{tab:gemma_example}
\begin{tabular}{lp{0.82\textwidth}}
\toprule
\textbf{Speaker} & \textbf{Utterance} \\
\midrule

Speaker\_1 &
\texthindi{मुझे नहीं लगता कि यह} airstrike \texthindi{इतनी} simple \texthindi{थी,} definitely \texthindi{कुछ} internal intelligence leak \texthindi{हुआ है।} Israel \texthindi{की} intelligence capabilities impressive \texthindi{हैं, पर} hideout \texthindi{का} exact location \texthindi{पता होना} suspicious \texthindi{लगता है। यह} whole situation \texthindi{अब बहुत} unpredictable \texthindi{हो गई है।} \\

Speaker\_2 &
Exactly, \texthindi{मुझे भी यही लगता है} because precise targeting \texthindi{बिना} inside information \texthindi{के} impossible \texthindi{है।} My cousin works in a security agency and he always says that human intelligence is the biggest weak point in any high-profile operation. \texthindi{यह} incident prove \texthindi{करता है कि} Israel \texthindi{का} network Iran \texthindi{के} inside \texthindi{बहुत} deep \texthindi{है।} Do you think Iran will now launch a massive counter-attack? \\

Speaker\_1 &
\texthindi{मेरे एक} friend \texthindi{ने} recently \texthindi{एक} security conference attend \texthindi{की थी, जहाँ उन्होंने} discuss \texthindi{किया था कि कैसे} high-ranking officials \texthindi{की} digital footprints \texthindi{उन्हें} vulnerable \texthindi{बनाती हैं।} He mentioned \texthindi{कि} modern warfare \texthindi{में} physical walls \texthindi{से ज़्यादा} cybersecurity important \texthindi{हो गई है। शायद} Larijani \texthindi{की} location \texthindi{भी किसी} encrypted communication leak \texthindi{की वजह से} expose \texthindi{हुई होगी।} \\

Speaker\_2 &
\texthindi{यह} point \texthindi{एकदम} valid \texthindi{है} because digital leaks \texthindi{आजकल} common \texthindi{हो गए हैं।} My old college roommate used to work in a tech firm and he once showed me how easily public Wi-Fi can expose a person's real-time location. \texthindi{यह देखकर मैं} shock \texthindi{हो गया था कि} basic mistakes \texthindi{कितनी} dangerous \texthindi{हो सकती हैं।क्या आपको लगता है कि} Iran \texthindi{अपनी} security protocols \texthindi{को} completely change \texthindi{करेगा?} \\

Speaker\_1 &
\texthindi{मुझे नहीं लगता कि वे} security protocols \texthindi{को} completely change \texthindi{करेंगे,} because it's almost impossible to overhaul an entire system overnight. \texthindi{वे शायद सिर्फ} high-level officials \texthindi{के} communication channels \texthindi{को} narrow \texthindi{कर देंगे ताकि} leaks \texthindi{कम हों।} system change \texthindi{करने के बजाय वे} revenge \texthindi{लेने पर ज़्यादा} focus \texthindi{करेंगे।} \\

Speaker\_2 &
I disagree with this perspective because only revenge \texthindi{लेने से} problem solve \texthindi{नहीं होगी।} Iran \texthindi{को} realize \texthindi{करना होगा कि} their current security infrastructure is fundamentally broken and needs a complete overhaul. \texthindi{अगर वे} sirf communication narrow \texthindi{करेंगे, तो} Israel’s intelligence network still find a way to penetrate their system Israel's intelligence network \texthindi{क्या आपको नहीं लगता कि} without a systemic change, \texthindi{वे और भी ज़्यादा} vulnerable \texthindi{रहेंगे?} \\

\bottomrule
\end{tabular}
\end{table*}

\begin{table*}[t]
\centering
\small
\caption{Example Hindi-English code-mixed conversation generated by the proposed pipeline- model GPT-OSS-120B.}
\label{tab:gptoss_example}
\begin{tabular}{lp{0.82\textwidth}}
\toprule
\textbf{Speaker} & \textbf{Utterance} \\
\midrule

Speaker\_1 &
\texthindi{यह} news \texthindi{सुनकर मुझे बहुत} shock \texthindi{लगा कि} Israel \texthindi{ने सीधे} Iran \texthindi{के सबसे} senior political figure, Ali Larijani, \texthindi{को} airstrike \texthindi{से मार दिया, और साथ में} Basij commander Gholamreza Soleimani \texthindi{भी} eliminated \texthindi{बताया गया । इस तरह का} high-profile targeted killing \texthindi{पहले कभी नहीं हुआ, और ये} Iran \texthindi{की} national security council \texthindi{ने भी} confirm \texthindi{किया है, जो} war \texthindi{के} third week \texthindi{में पहले ही} 2,000+ casualties \texthindi{के बीच है। अब} IRGC \texthindi{की} political influence \texthindi{में बड़ा} void create \texthindi{हो गया है, और} regional dynamics \texthindi{पूरी तरह} shift \texthindi{हो सकते हैं ।} \\

Speaker\_2 &
\texthindi{मैं समझता हूँ आपका} shock, \texthindi{लेकिन यह कहना सही नहीं कि ऐसा पहले कभी नहीं हुआ; पिछले साल में भी कई} senior Iranian officials \texthindi{को} targeted killings \texthindi{का सामना करना पड़ा था। साथ ही,} IRGC \texthindi{की} political influence \texthindi{में पूरी तरह} void \texthindi{नहीं बनेगा, क्योंकि उनका} institutional network \texthindi{बहुत} deep \texthindi{और} resilient \texthindi{है। फिर भी, इस नए} void \texthindi{से} regional dynamics \texthindi{में ज़रूर कुछ बदलाव आने की संभावना है।} \\

Speaker\_1 &
\texthindi{ये बात सुनते ही दिमाग में ये सवाल उठ रहा है कि क्या} Israel \texthindi{अब पूरे} Middle-East \texthindi{में ऐसे} high-value targets \texthindi{को एक-एक करके} eliminate \texthindi{कर रहा है? अगर ऐसा है तो} IRGC \texthindi{की} internal hierarchy \texthindi{में कौन-से नए} power-centers \texthindi{उभर सकते हैं, इसका अंदाज़ा लगाना मुश्किल है। साथ ही,} Hormuz Strait \texthindi{की बंदिशें कितनी देर तक टिकेंगी, इस पर भी बहुत} curiosity \texthindi{है, क्योंकि} global oil flow \texthindi{में इसका असर बहुत बड़ा होगा।} \\

Speaker\_2 &
\texthindi{मैं नहीं मानता कि} Israel \texthindi{अब पूरे} Middle-East \texthindi{में हर} high-value target \texthindi{को} systematic \texthindi{तरीके से} eliminate \texthindi{कर रहा है, क्योंकि ऐसी} strategy \texthindi{उनका} operational bandwidth \texthindi{और} diplomatic cost \texthindi{बहुत बढ़ा देगी। वास्तव में, वे अक्सर} specific intelligence-driven missions \texthindi{पर} focus \texthindi{करते हैं, जहाँ} political payoff \texthindi{सबसे ज़्यादा हो, न कि व्यापक} regional sweep \texthindi{।} Hormuz Strait \texthindi{की बंदिशें भी शायद कुछ हफ़्तों तक ही रहेंगी, क्योंकि} global oil market \texthindi{का} pressure \texthindi{दोनों पक्षों को जल्दी ही} negotiation \texthindi{की दहलीज़ पर ले आएगा ।} \\

Speaker\_1 &
\texthindi{क्या} Israel \texthindi{अगले हफ़्ते तक किसी और} high-value target \texthindi{को भी} pinpoint \texthindi{कर लेगा, या ये सिर्फ एक} strategic message \texthindi{है?अगर} IRGC \texthindi{के अंदर नए} power-centers \texthindi{उभरते हैं, तो उनका कौन-सा} faction \texthindi{सबसे ज़्यादा} influence \texthindi{हासिल करेगा, इस पर भी बहुत} curiosity \texthindi{है। इस बीच} Hormuz \texthindi{की बंदिशें कब तक चलेंगी, और} oil markets \texthindi{में इसका} real-time impact \texthindi{कैसे दिखेगा, देखना काफ़ी} interesting \texthindi{रहेगा।} \\

Speaker\_2 &
\texthindi{मैं मानता हूँ कि यह सिर्फ एक संदेश नहीं, बल्कि एक} calculated move \texthindi{है, लेकिन मैं इस बात से असहमत हूँ कि} Israel \texthindi{अगले हफ़्ते तक नया} high-value target \texthindi{ले लेगा; उनके} operational tempo \texthindi{और} diplomatic backlash \texthindi{अक्सर उन्हें ऐसा जल्दी करने से रोकते हैं।} IRGC \texthindi{में कौन-सा} faction \texthindi{हावी होगा, इसका अनुमान अभी} premature \texthindi{है क्योंकि कई} parallel networks \texthindi{अभी भी सक्रिय हैं ।} Hormuz \texthindi{की बंदिशें संभवतः कुछ हफ़्तों में ही ढीली होंगी, क्योंकि} oil market \texthindi{की} volatility \texthindi{दोनों पक्षों को जल्दी} pressure \texthindi{में लाती है।} \\

\bottomrule
\end{tabular}
\end{table*}

\end{document}